\documentclass[10pt, a4paper]{article}

\usepackage[]{lrec-coling2024}
\usepackage{rotating} 
\usepackage{xcolor}
\usepackage{colortbl}
\usepackage{booktabs}
\usepackage{multirow,multicol}
\usepackage{makecell} 
\usepackage{graphicx}
\usepackage{caption}
\usepackage{subcaption}
\usepackage{enumitem}
\usepackage{listings}
\usepackage{amssymb}
\usepackage{pifont}
\newcommand{\cmark}{\ding{52}}
\newcommand{\xmark}{\ding{56}}

\title{Recent Trends in Personalized Dialogue Generation: \\A Review of Datasets, Methodologies, and Evaluations}

\name{Yi-Pei Chen*$\dagger$, Noriki Nishida$\dagger$, Hideki Nakayama*, Yuji Matsumoto$\dagger$} 

\address{* University of Tokyo, $\dagger$ RIKEN AIP\\
         ypc@g.ecc.u-tokyo.ac.jp, nakayama@ci.i.u-tokyo.ac.jp\\
         \{noriki.nishida, yuji.matsumoto\}@riken.jp\\}

\abstract{
Enhancing user engagement through personalization in conversational agents has gained significance, especially with the advent of large language models that generate fluent responses. Personalized dialogue generation, however, is multifaceted and varies in its definition -- ranging from instilling a persona in the agent to capturing users' explicit and implicit cues. This paper seeks to systemically survey the recent landscape of personalized dialogue generation, including the datasets employed, methodologies developed, and evaluation metrics applied. Covering 22 datasets, we highlight benchmark datasets and newer ones enriched with additional features. We further analyze 17 seminal works from top conferences between 2021-2023 and identify five distinct types of problems. We also shed light on recent progress by LLMs in personalized dialogue generation. Our evaluation section offers a comprehensive summary of assessment facets and metrics utilized in these works. In conclusion, we discuss prevailing challenges and envision prospect directions for future research in personalized dialogue generation.
 \\ \newline  }

\begin{document}

\maketitleabstract

\section{Introduction}
Personalization can enhance a user's engagement with conversational agents \cite{zhang-etal-2018-personalizing,kwon-etal-2023-ground}. The ability of large language models (LLMs) \cite{GPT4,llama2} to generate fluent and coherent responses to human queries underscores the importance of building personalized systems that cater to each individual's background and preferences more than ever before.

\begin{figure}
    \centering
    \includegraphics[trim={0.8cm 1 0 0},clip,width=1\columnwidth]{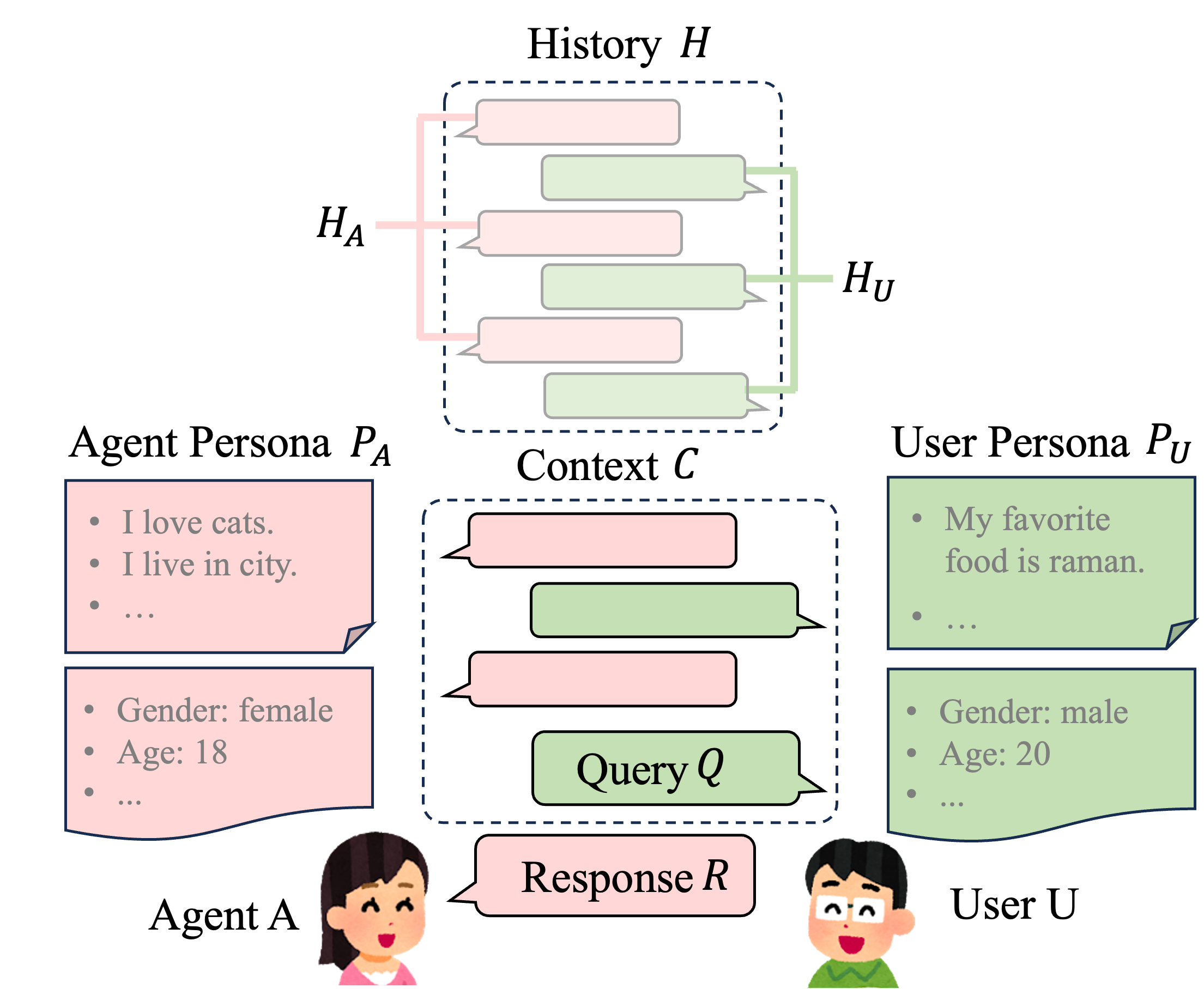}
    \caption{An overview of personalized dialogue generation. Assumed that the conversation is performed by two speakers, i.e., an agent $A$ and a user $U$, the goal is to generate the response $R$ given the dialogue context $C$ or the last utterance $Q$, plus the persona of the agent or user ($P_A$ or $P_U$) (explicit), or utterance histories of them ($H_A$ or $H_U$) (implicit).}
    \label{fig:problem_definition}
\end{figure}

However, \textit{personalization} remains an open question, with varying definitions among different individuals. Figure~\ref{fig:problem_definition} illustrates the three scenarios currently being explored in research on personalized dialogue generation. This may involve endowing the agent with a persona, modeling the other party's persona, or both. In this context, a persona refers to characteristics such as personal background, interests, or behaviors that shape the identity or personality of the user or the agent.
Personalized response generation can be seen as a conditional text generation task where a response is generated based on the given context and conditioned on either the speakers' explicitly provided persona or implicit attributes embedded in the dialogue history.

\begin{table*}[!htp]
\centering
\scriptsize
\resizebox{1.0\linewidth}{!}{
\begin{tabular}{lllllrrccccl}
\toprule
\textbf{Dataset} & & \textbf{Lang} &\textbf{Persona} &\textbf{Source} &\makecell[c]{\textbf{D}} & \textbf{U/D} &\textbf{Role} & \textbf{Pub.} & \makecell[c]{\textbf{P} \\ \textbf{Ground.}} & \makecell[c]{\textbf{Multi}\\ \textbf{Session}} &\textbf{Feature} \\
\midrule
PersonaChat &\citeyearpar{zhang-etal-2018-personalizing} &EN &Description &Crowd &10.9K &\textbf{14.9} &H-H &\cmark &\xmark &\xmark & \\
ConvAI2 &\citeyearpar{dinan2020second} &EN &Description &Crowd &11.0K &\textbf{15.0} &H-H &\cmark &\xmark &\xmark & \\
BlendedSkillTalk &\citeyearpar{smith-etal-2020-put} &EN &Description &Crowd &6.8K &\textbf{11.2} &H-H &\cmark &\xmark &\xmark &KG, EMP \\
MSC &\citeyearpar{xu-etal-2022-beyond} &EN &Description &Crowd &24.0K &\textbf{12.5} &H-H &\cmark &\xmark &\cellcolor[HTML]{d9ead3}\cmark & \\
FoCus &\citeyearpar{jang2022call} &EN &Description &Crowd &13.5K &\textbf{11.3} &\cellcolor[HTML]{d9d9d9}H-A &\cmark &\cellcolor[HTML]{d9ead3}\cmark &\xmark &KG, ConvQA \\
PersonaMinEdit &\citeyearpar{wu-etal-2021-transferable} &EN &Description &Crowd &121.8K &2.0 &H-H &\cmark &\xmark &\xmark &OOD \\
IT-ConvAI2 &\citeyearpar{liu2022improving} &EN &Description &ConvAI2 &1.6K &2.0 &H-H &\cmark &\xmark &\xmark &OOD \\
Reddit &\citeyearpar{mazare-etal-2018-training} &EN &Description &Reddit &\textbf{700M} &- &H-H &\cellcolor[HTML]{f4cccc}\xmark &\xmark &\xmark & \\
PEC &\citeyearpar{zhong-etal-2020-towards} &EN &Description &Reddit (H, O) &355.0K &2.4 &H-H &\cmark &\xmark &\xmark &EMP \\
MPChat &\citeyearpar{ahn-etal-2023-mpchat} &EN &Description &Reddit (648*) &15.0K &2.8 &H-H &\cmark &\cellcolor[HTML]{d9ead3}\cmark &\xmark &Multi-modal \\
PER-CHAT &\citeyearpar{wu-etal-2021-personalized} &EN &KV + History &Reddit (A) &\textbf{1.5M} &2.0 &H-H &\cmark &\xmark &\xmark &QA \\
Reddit (DialoGPT) &\citeyearpar{zhang-etal-2020-dialogpt} &EN &UserID &Reddit &\textbf{147M} &- &H-H &\cmark &\xmark &\xmark & \\
Persona Reddit &\citeyearpar{zeng-nie-2021-simple} &EN &UserID &Reddit &\textbf{3.1M} &- &H-H &$\triangle$ &\xmark &\xmark & \\
\midrule
Baidu PersonaChat &N/A &ZH &Description &Crowd &24.5K &\textbf{16.3} &H-H &\cmark &\xmark &\xmark & \\
DuLeMon &\citeyearpar{xu-etal-2022-long} &ZH &Description &Crowd &27.5K &\textbf{16.3} &\cellcolor[HTML]{d9d9d9}H-A &\cmark &\cellcolor[HTML]{d9ead3}\cmark &\cellcolor[HTML]{d9ead3}\cmark & \\
LiveChat &\citeyearpar{gao-etal-2023-livechat} &ZH &Description + KV &Douyin &1.3K &2.0 &H-H &\cmark &\xmark &\xmark &LiveStream \\
WD-PB &\citeyearpar{ijcai2018p595} &ZH &KV &Weibo &76.9K &2.0 &H-H &$\triangle$ &\xmark &\xmark & \\
Personal Dialog &\citeyearpar{zheng2019personalized} &ZH &KV &Weibo &\textbf{20.8M} &2.7 &H-H &\cmark &\xmark &\xmark & \\
PchatbotW &\citeyearpar{qian2021pchatbot} &ZH &UserID &Weibo &\textbf{139.4M} &2.0 &H-H &$\triangle$ &\xmark &\xmark & \\
PchatbotL &\citeyearpar{qian2021pchatbot} &ZH &UserID &Forums &\textbf{59.4M} &2.0 &H-H &$\triangle$ &\xmark &\xmark & Judiciary \\ 
\midrule
MSPD &\citeyearpar{kwon-etal-2023-ground} &KR &Description &Crowd &53.9K &\textbf{11.2} &\cellcolor[HTML]{d9d9d9}H-A &\cellcolor[HTML]{f4cccc}\xmark &\cellcolor[HTML]{d9ead3}\cmark &\cellcolor[HTML]{d9ead3}\cmark & \\
\midrule
XPersona &\citeyearpar{lin2021xpersona} & Multi &Description &ConvAI2 & 3.3K & 15.6 & H-H & \cmark & \xmark & \xmark \\
\bottomrule
\end{tabular}
}
\caption{Dataset summary.
Persona can be represented by descriptive sentences (Description) or key-value dictionary (KV).
Data sources are mainly from crowdsourcing (Crowd) and Reddit (subreddit) (
H: happy,
O: offmychest,
A: AskReddit,
648*: from 648 subreddits, see the paper for the list).
D: number of dialogues.
U/D: average utterance per dialogue.
H-H: human-human conversation.
H-A: human-agent conversation.
$\triangle$: Available upon request.
Persona Grounding (P Ground) means each utterance has the label of which persona sentence the utterance is grounding on.
KG: knowledge.
EMP: empathy.
OOD: out-of-distribution persona. We added background colors to emphasize the less-frequent options.}
\label{tab:dataset_summary}
\end{table*}
We aim to find out what is being personalized (\textbf{Dataset}), how the dialogue systems implement personalization (\textbf{Methodology}), and how previous research evaluates the personalization (\textbf{Evaluation}) in this systematic survey.
In the Dataset Section (Sec.~\ref{sec:dataset}), we introduce 22 datasets, including benchmark datasets that were frequently used in previous personalized dialogue research and recently published datasets that propose more features to add to the existing ones. We summarize them in Table~\ref{tab:dataset_summary}.
In the Methodology Section (Sec.~\ref{sec:approach}), we center on 17 works published in the three years, from 2021 to Oct. 2023, at top conferences, including ACL, NAACL, EMNLP, AAAI, etc., based on keyword searching and the related works in each paper.
In the Evaluation Section (Sec.~\ref{sec:eval}), we summarize each work's evaluated aspects and evaluation metrics.

We address the challenges and potential future trajectories in terms of Datasets, Methodologies, and Evaluation in the Discussion Section (Sec.~\ref{sec:discuss}).
We believe that the primary issues with current datasets revolve around their size, quality, and diversity. In terms of methodology, we stress the need to examine the assumptions underpinning approaches to personalization critically. Finally, we advocate for a standardized evaluation benchmark equipped with advanced metrics to assess contributions in this domain fairly.

\section{Datasets}
\label{sec:dataset}
In this section, we first review datasets that have been used in personalized dialogue generation literature.
We then discuss the characteristics of the datasets, with a focus on persona representations and domain and language biases.

\begin{table*}[]
\centering
\small
\resizebox{1.0\linewidth}{!}{
\begin{tabular}{l|l}
    Dataset & Attribute Keys \\
    \midrule
    WD-PB \cite{ijcai2018p595} & gender, location, age, name, weight, constellation \\
    Personal Dialog \cite{zheng2019personalized} & gender, location, age, self-description, interest tags \\
    PER-CHAT \cite{wu-etal-2021-personalized} & gender, location, self-description, ID, pets, family, favorites, partner, possessions \\
    LiveChat \cite{gao-etal-2023-livechat} & gender, location, age, ID, character, fans number, live time, reply barrage, audiences, skill \\ 
\end{tabular}
}
\caption{The types of personal attributes collected in each key-value dataset.}
\label{tab:datasets_key-value}
\end{table*}
\subsection{Datasets Review}
Table~\ref{tab:dataset_summary} summarizes the datasets.
One of the first and most widely used dataset in personalized dialogue research is the \textbf{PersonaChat} dataset \cite{zhang-etal-2018-personalizing}, which is also used in The Second Conversational Intelligence Challenge (ConvAI2) \cite{dinan2020second}.
Nine out of eighteen works introduced in Sec.~\ref{sec:approach} evaluated their systems on PersonaChat/ConvAI2. 
PersonaChat consists of 10.9K English dialogues in total, which were collected through crowdsourcing.
Each dialogue in PersonaChat consists of two human speakers trying to know each other, and about five persona descriptive sentences are provided to each speaker.

The \textbf{FoCus} dataset \cite{jang2022call} introduced \textit{persona grounding}, indicating the specific persona sentence to which each utterance is anchored.
With the persona grounding label, models can learn to extract the most relevant personal information from persona.
Datasets such as \textbf{MPChat}~\cite{ahn-etal-2023-mpchat}, \textbf{DuLeMon}~\cite{xu-etal-2022-long}, and \textbf{MSPD}~\cite{kwon-etal-2023-ground} also contain persona grounding.

Some datasets introduce multi-session dialogues.
The \textbf{MSC} dataset \cite{xu-etal-2022-beyond} is similar to PersonaChat, except that MSC has four to five sessions for the same pair of speakers. 
\textbf{DuLeMon} \cite{xu-etal-2022-long} and \textbf{MSPD} \cite{kwon-etal-2023-ground} also introduced multi-session into their datasets, where the former is in Chinese and the latter is in Korean.

Some datasets also augment the persona with other features such as knowledge, empathy, and vision.
The \textbf{BlendedSkillTalk} (BST) dataset \cite{smith-etal-2020-put} differs from PersonaChat in that BST endows agents persona, knowledge (KG), and empathy (EMP) by combining PersonaChat with Wizard of Wikipedia \cite{dinan2019wizard} and Empathetic Dialogues \cite{rashkin2019towards}. 
Each agent is given two persona sentences and each utterance is labeled with the type of bias they were grounded on (persona, KG, or EMP).
The \textbf{Persona-based Empathetic Conversation} (PEC) dataset \cite{zhong-etal-2020-towards} also investigated the impact of persona on empathetic responding.
PEC differs from BST in extracting personas from Reddit.
The \textbf{MPChat} dataset \cite{ahn-etal-2023-mpchat} introduced the first multi-modal persona, where the persona is not only text descriptions but also paired with an image.

Due to the limited scale of crowdsourced datasets, the persona distribution in real-world data often exceeds that of the datasets. 
\textbf{PersonaMinEdit} \cite{wu-etal-2021-transferable} and \textbf{Inadequate-Tiny-ConvAI2} (IT-ConvAI2) \cite{liu2022improving} are specifically designed to test the generation grounding on unseen personas. 
Specifically, these datasets ensure that their test set contains only unseen personas the response should conditioned on by either rewriting or filtering.

\subsection{Facets}
\subsubsection{Persona Representation}
It is not obvious how the persona should be represented.
Through the literature survey, we found that different datasets employ different ways of representing persona information and that the persona representation can be classified into three categories: (1) persona description, (2) key-value attributes, (3) user ID and comment histories.

Most datasets employ \textbf{descriptive sentences} as the persona representation \cite{dinan2020second,mazare-etal-2018-training,smith-etal-2020-put,zhong-etal-2020-towards,wu-etal-2021-transferable,xu-etal-2022-beyond,xu-etal-2022-long,liu2022improving,jang2022call,ahn-etal-2023-mpchat,kwon-etal-2023-ground}. 
For example, PersonaChat~\cite{zhang-etal-2018-personalizing} contains 5 descriptive sentences for each speaker.
These datasets primarily recruited annotators to chat based on given persona descriptions, thus avoiding privacy concerns.
\citet{mazare-etal-2018-training} extracted persona descriptions from Reddit using heuristic rules to gather large datasets, which was followed by \citet{zhong-etal-2020-towards} and \citet{ahn-etal-2023-mpchat}. 

Some datasets represent personal information using \textbf{sparse key-value attributes} \cite{ijcai2018p595,zheng2019personalized,wu-etal-2021-personalized,gao-etal-2023-livechat}. 
Table~\ref{tab:datasets_key-value} shows the examples of key-value attributes.
For example, WD-PB \cite{ijcai2018p595} defines 6 attribute keys like gender, location, and age, and the values corresponding to these keys are recorded as persona information for each user.
In addition to key-value attributes, LiveChat \cite{gao-etal-2023-livechat} provides dense persona description extracted by a rule-based method similar to \citet{mazare-etal-2018-training}, and PER-CHAT \cite{wu-etal-2021-personalized} provides query-related comment histories extracted by a pretrained IR system as additional personal information.
They first define what attribute types they want to extract, then collect the results from posts/responses using regular expression \cite{ijcai2018p595,wu-etal-2021-personalized}, from the profile provided by the users \cite{zheng2019personalized, gao-etal-2023-livechat}, or recruit annotators to label from the given context \cite{gao-etal-2023-livechat}.

Some datasets collected from social platforms only provide \textbf{speaker/user IDs} as persona information \cite{zhang-etal-2020-dialogpt,qian2021pchatbot,zeng-nie-2021-simple}.
They assume that the speaker IDs are used to retrieve comment histories of the corresponding users in the social platform.
The speaker IDs have also been used in other tasks, such as speaker identification.
They consider the same user's utterances to contain implicit personal information, and personalization can be measured by the similarity between the generated response and the ground-truth response.

\subsubsection{Domain and Language Biases}
\label{sec:dataset-domainNlang}
Personalized dialogue generation is an open-domain task:
Human speakers are allowed to talk about whatever topics they like.
However, through the literature survey, we found that there are domain differences between datasets.
PchatbotL \cite{qian2021pchatbot} was collected from Chinese judicial forums.
PER-CHAT \cite{wu-etal-2021-personalized} crawled from Subreddit AskMeQuestion is more similar to a question answering (QA) dataset.
FoCus \cite{jang2022call} asked annotators to discuss landmarks from Google Landmarks Dataset v2 \cite{weyand2020google}. It is more like a conversational question answering dataset where a user asks questions about a landmark and the agent answers, rather than a natural conversation between humans.
LiveChat \cite{gao-etal-2023-livechat} gathered data from live streaming on Douyin (Chinese TikTok). There might be multiple people (multi-party) responding to the streamer, while the streamer only responds to one, making it a 1-1 dialogue. 

\begin{table*}
\centering
\small
\resizebox{1.0\linewidth}{!}{
\begin{tabular}{lllllllll}
\toprule
\multicolumn{2}{c}{\multirow{2}{*}{\textbf{Model}}} &\multirow{2}{*}{\makecell[c]{\textbf{Whose}\\\textbf{Persona}}} &\multirow{2}{*}{\textbf{Approaches}} &\multirow{2}{*}{\textbf{Dataset}} &\multicolumn{3}{c}{\textbf{Model Input}} &\multirow{2}{*}{\makecell[c]{\textbf{NLI}}} \\\cmidrule{6-8}
& & & & & $\mathbf{D}$ & $\textbf{P}$ &\textbf{Other} & \\\midrule
UA-CVAE &\citeyearpar{lee2022improving} &Self &Coherence &CAI, ED &$C$ &$P_A$ & & \\
BoB &\citeyearpar{song-etal-2021-bob} &Self &Consistency &CAI, PD &$Q$ &$P_A$ & &\cmark \\
PCF &\citeyearpar{wang2023please} &Self &Consistency, Coherence &CAI, PD &$Q$ &$P_A$ & &\cmark \\
LMEDR &\citeyearpar{chen2023learning} &Self &Consistency, Coherence &PC, AVSD &$Q$ &$P_A$ & &\cmark \\
SimOAP &\citeyearpar{zhou-etal-2023-simoap} &Self &Consistency, Coherence &PC &$C$ &$P_A$ & &\cmark \\
PAA &\citeyearpar{huang2023personalized} &Self &Balance &CAI &$C$ &$P_A$ & & \\
D3 &\citeyearpar{cao-etal-2022-model} &Self &Data Scarcity &PC &- &- & &\cmark \\
GME &\citeyearpar{wu-etal-2021-transferable} &Self &Long-tail &PME, BST &$Q$ &($P_A$) &$P_{\mathsf{OOD}}$ & \\
PS-Transformer &\citeyearpar{liu2022improving} &Self &Select , OOD, Long-tail &CAI, IT-CAI &$Q$ &$P_A$ &$P_{\mathsf{OOD}}$, P Pool &\cmark \\
DHAP &\citeyearpar{ma2021one} &Self &Unknown &PCW, Reddit-D &$Q$ &$H_A$ & & \\
\midrule
FoCus &\citeyearpar{jang2022call} &Other &Select &FoCus &$C$ &$P_U$ &Knowledge & \\
INFO &\citeyearpar{lim-etal-2022-truly} &Other &Select &FoCus &$C$ &$P_U$ &Knowledge & \\
WWH &\citeyearpar{kwon-etal-2023-ground} &Other &Select, Balance &MSPD &$C$ &$P_U$ & & \\
IUPD &\citeyearpar{cho-etal-2022-personalized} &Other &Unknown &CAI &$C$ &($P_U$) & & \\
CLV &\citeyearpar{tang-etal-2023-enhancing-personalized} &Other &Unknown &CAI, Baidu PC &$Q$ &($P_U$) & & \\
MSP &\citeyearpar{zhong-etal-2022-less} &Other &Unknown &PCW, Reddit-D &$Q$ &$H_U$ &$H_{sim}$ & \\
\midrule
DuLeMon &\citeyearpar{xu-etal-2022-long} &Both &Select, Unknown &DuLeMon &$C$ &$H_A$, $H_U$ & & \\
\bottomrule
\end{tabular}
}
\caption{An overview of Personalized Dialogue Systems reviewed in Sec~\ref{sec:approach}.
We show
(1) \textbf{whose persona} each model aims to represent (Self: agent; Other: user),
(2) \textbf{approaches} involved in each work (Sec.~\ref{sec:approach-task}),
(3) \textbf{datasets} they are trained on
(PC: PersonaChat;
CAI: ConvAI2;
PD: PeresonalDialog;
PCW: PChatbotW;
Reddit-D: Reddit (DialogGPT);
ED: EmpatheticDialog;
PME: PersonaMinEdit;
BST: BlendedSkillTalk),
(4) \textbf{model inputs}, i.e., dialogue types D ($C$: context; $Q$: query), persona types P ($P_A$: agent persona; $P_U$: user persona; $H_A$: agent history; $H_U$: user history), other modalities ($P_{\mathsf{OOD}}$: out-of-distribution persona; P Pool: additional personas pool),
(5) and whether additional \textbf{NLI} data/model is used.
The parentheses like ``($P_A$)'' indicate that the persona information is only provided during training.
}
\label{tab:model_summary}
\end{table*}
Also, persona dialogue generation is not limited to specific languages;
however, the languages of the datasets are highly biased.
XPersona (Multilingual Persona-Chat) \cite{lin2021xpersona} translates a portion of PersonaChat into Chinese, French, Indonesian, Italian, Korean, and Japanese. However, the number of dialogues in each language is very limited. For example, there are only 280 dialogues in Italian. 
There are other translations of PersonaChat, such as Japanese \cite{sugiyama2021empirical} and Korean\footnote{\url{https://aihub.or.kr/}}.
Besides translation, Baidu constructed and released the Chinese PersonaChat dataset \footnote{\url{https://www.luge.ai/\#/luge/dataDetail?id=38}} which is similar to PersonaChat.
DuLeMon \cite{xu-etal-2022-long} and Multi-Session Personalized Dialogue (MSPD) \cite{kwon-etal-2023-ground} are the Chinese and Korean version of Multi-Session Chat (MSC) \cite{xu-etal-2022-beyond}.
Beyond the MSC dataset, both DuLeMon and MSPD have additional persona grounding labels that explicitly show which persona sentence the utterance is grounding on.

\section{Methodology}
\label{sec:approach}
In this section, we first introduce the task definition of personalized dialogue generation, and then discuss recent methodology advances in personalized dialogue generation published at top conferences 
from 2021 to 2023.
Finally, we review recent studies on large language models in personalized dialogue generation.

\subsection{Problem Statement}
\label{sec:approach-statement}

In this survey, we focus on the bilateral conversation, i.e., the conversation between two parties.
As shown in Fig.~\ref{fig:problem_definition}, we define two speakers as an agent $A$ and a user $U$ with the current dialogue context $C$, where the last utterance of $C$ is defined as the query $Q$ uttered by the user $U$. 
There might also be past dialogue history $H$, where the history utterances by the agent $A$ and the user $U$ are denoted as $H_A$ and $H_U$, respectively.

Personalized dialogue generation aims to generate a response $R$ as an agent conditioned on the input dialogue $D$ and the persona $P$.
The input dialogue $D$ can be a single query $Q$ (i.e., the last utterance by the user) or the dialogue context $C$.
The persona $P$ can be in various representation formats, such as descriptive sentences and sparse key-value attributes, as seen in Section~\ref{sec:dataset}.
Persona $P$ for agent $A$ is denoted as $P_A$, while that for user $U$ is represented as $P_U$.

There are two main streams of personalized dialogue generation.
One direction endows the agent with its own persona $P_A$ and focuses on generating $R$ consistent with persona $P_A$ and coherent with the dialogue context $C$.
In most studies, $P_A$ is explicitly provided as an input;
one exception is \citet{ma2021one}, in which $P_A$ is not provided and the agent's history responses $H_A$ are considered as the persona information.

The other direction aims to model the user's persona $P_U$ to generate responses that are better tailored to the user's needs. The emphasis is often on selecting segments from the provided $P_U$ that are most relevant to $Q$, or on establishing $P_U$ when it is not directly given. In cases where $P_U$ is absent, one might derive explicit or implicit $P_U$ from the user's dialogue history $H_U$, or infer implicit $P_U$ via conditional variational inference.

\subsection{Approaches}
\label{sec:approach-task}
Through the survey, we found that recent approaches to personalized dialogue generation can be classified into 5 groups based on their motivations and target issues: Consistency and Coherence, Persona-Context Balancing, Relevant Persona Selection, Unknown Persona Modeling, and Data Scarcity.
Note that a single paper may cover more than one of these issues.

As a side note, although a closely related line of research focuses on extracting personas from dialogues, such as \citet{zhu-etal-2023-paed}, it falls beyond the scope of this paper, which is concentrated on personalized dialogue generation.

\subsubsection{Consistency and Coherence}
\label{sec:approach-task-consistNcohere}
As mentioned in the previous section, most research on endowing models with personas focuses on generating responses that are simultaneously consistent with the given persona and coherent with the context.

Uncertainty Aware CVAE (UA-CVAE) \cite{lee2022improving} tackled the coherence problem using conditional variational autoencoder (VAE) training. They proposed to generate response $R$ conditioned on the context $C$, the agent persona $P_A$, and the latent variable $z$. The latent variable $z$ is sampled from latent Gaussian distribution $p(z|C, P_A)$, where the variance of $z$ acts as an approximation to the uncertainty in the input ($C$, $P_A$).

Natural Language Inference (NLI) is commonly used to solve the consistency problem, which predicts whether a premise and a hypothesis are entailed, neutral, or contradictory.
BoB \cite{song-etal-2021-bob} trained a decoder with NLI data to ensure consistency and minimize contradiction between the response $R$ and the agent persona $P_A$. 

Based on BoB, PCF \cite{wang2023please} further added another NLI module between $R$ and the query $Q$ to maintain the coherence of the dialogue. 
LMEDR \cite{chen2023learning} fine-tuned a pretrained NLI model with two additional matrix parameters that act as additional ``memory'' for language modeling, one for consistency and the other for coherence.
SimOAP \cite{zhou-etal-2023-simoap} demonstrated that responses with high probabilities aren't always superior to those with lower probabilities. Consequently, their approach involves generating an extensive list of response candidates and post-filtering. The candidates are first filtered for coherence with $C$ using the TF-IDF method \cite{salton1988term}, then selected for consistency with  $P_A$ through a pretrained NLI model.

\subsubsection{Persona-Context Balancing}
Since not all responses need personalization, deciding when to focus more on the context and when to weave more personal information into the response is important. 
Persona-Adaptive Attention (PAA) \cite{huang2023personalized} separately encode $P_A$ and $C$, then design attention mechanism to combine them dynamically. 
WWH \cite{kwon-etal-2023-ground} integrates non-personalized datasets with the personalized dataset, adjusting training data sampling based on each dataset's size to yield more natural responses.

\subsubsection{Relevant Persona Selection}
While not all information in the given persona $P \in \{P_A, P_U\}$ is related to the dialogue $D \in \{Q, C\}$, selecting the most relevant persona sentence becomes crucial to generate a natural and engaging response.
PS-Transformer \cite{liu2022improving} and FoCUS \cite{jang2022call} trained a binary classifier for each persona sentence to evaluate their likelihood of being utilized. 
In contrast, INFO \cite{lim-etal-2022-truly} obtained the weights over all $P_A$ candidates via a multi-class classifier.
DuLeMon \cite{xu-etal-2022-long} and WWH \cite{kwon-etal-2023-ground} trained LLM to discriminate the negative persona sentences from the positive ones based on $C$.
Note that a dataset with persona-grounding labels - each utterance associated with a specific persona attribute - is usually required to learn relevant persona selection.

\subsubsection{Unknown Persona Modeling}
In the case when persona $P$ is not explicitly given, the personal information could be extracted from dialogue histories of the speaker (${H_A}$ or $H_U$) \cite{ma2021one,zhong-etal-2022-less,xu-etal-2022-long}, or implicitly modeled by latent variables \cite{cho-etal-2022-personalized,tang-etal-2023-enhancing-personalized}.

DHAP \cite{ma2021one}, MSP \cite{zhong-etal-2022-less}, and DuLeMon \cite{xu-etal-2022-long} extract explicit (DuLeMon, MSP) or implicit (DHAP) personal information from dialogue histories. 
DuLeMon was the first to conduct personal information management for \textbf{both} $U$ and $A$, training a classifier to determine whether a clause in an utterance contains personal information. Following a similarity check, clauses containing persona information were then added or updated in either $P_A$ or $P_U$.
MSP derived the user persona $P_U$ from the dialogue history of the user $H_U$ and similar users $H_{sim}$. Histories unrelated to the query $Q$ are filtered out. From the remaining histories, the top $k$ tokens are selected based on the attention weight between $Q$ and the histories.
DHAP encoded $H_A$ as implicit agent persona $P_A$ and constructed a personalized vocabulary consisting of words from $H_A$. During decoding, DHAP switched between this personalized vocabulary and a general one. 

Both IUPD \cite{cho-etal-2022-personalized} and CLV \cite{tang-etal-2023-enhancing-personalized} model implicit user persona from $D$ via conditional variational inference (CVAE) training. 
CVAE has been used in dialogue generation to address the challenge of producing diverse responses for a single query. It also facilitates response generation under various conditions \cite{NIPS2015_8d55a249,zhao-etal-2017-learning, ijcai2019p721, chen-etal-2022-dialogved}\footnote{Not included in this work because they are not specifically designed for personalized dialogue generation.}.

IUPD proposed both a persona latent variable $Z_P$ and a fader latent variable $Z_\alpha$, along with special tokens $tok_P$ and $tok_\alpha$ for the input. 
The $Z_P$ variable captures the latent distribution of $P_U$, thereby connecting the context $C$ to the response $R$. 
Meanwhile, $Z_\alpha$ measures the extent to which $Z_P$'s persona information impacts $R$ under $C$.
The response generation is expressed as:
$p(R | Z_P, Z_\alpha, C) = p(R | Z_P, Z_\alpha, C)  p(Z_P | C) p(Z_\alpha | Z_P, C)$, 
where $p(R | Z_P, Z_\alpha, C)$ is the generator and $p(Z_P | C)$, $p(Z_\alpha | Z_P, C)$ are prior networks.
For $Z_P$, inputs to its prior and recognition networks combine the variable token with the condition, namely $[tok_P, C]$ and $[tok_P, P]$. 
For $Z_\alpha$, these are $[tok_\alpha, Z_P, C]$ and $[Z_\alpha, P, R]$.

CLV also proposed a persona latent variable $Z_P$ and additionally, a response latent variable $Z_R$.
They assume that $Z_P$ and $Z_R$ are independent. 
The response generation is formulated as:
$p(R | Z_P, Z_R, C) = p(R | Z_P, Z_R, C)  p(Z_P | C) p(Z_R | C)$.
In contrast to IUPD, CLV's prior and recognition networks for $Z_P$ utilize $Q$ and $[Q, P]$. For $Z_R$, the inputs are $Q$ and $[Q, R]$.

Note that although UA-CVAE \cite{lee2022improving} also using CVAE, they have the persona $P_A$ as input during inference time and is described in Sec.~\ref{sec:approach-task-consistNcohere}.

\subsubsection{Data Scarcity}
\label{sec:approach-task-datascarcity}
As shown in the Dataset section (Sec.~\ref{sec:dataset}), crowdsourced persona-based datasets usually have more dialogue context but are notable for their limited size.
Data augmentation is a straightforward answer to solve the data scarcity problem.
D$^3$ \cite{cao-etal-2022-model} is a model-agnostic method that purely manipulates the data. 
They kept only persona-related ($Q$, $R$) dialogues, removed irrelevant $P_A$ which are not entailed by $R$, and enlarged the number of persona-related dialogues to 1.8 times and personas to 3 times by BERT, GPT2, and back translation technique \cite{sennrich-etal-2016-improving}.
Although not trained with NLI data, pretrained NLI models have been extensively used to judge the consistency between the augmented $\tilde{P_A}$ and $\tilde{R}$, as well as to evaluate the coherence between the augmented $\tilde{Q}$ and $\tilde{R}$. 

Data scarcity also evokes the problem of out-of-distribution (OOD) personas. That is, the limited data provides the agent with a restricted persona $P_A$ and as a result, personas related to certain real-world queries might not be present in $P_A$ \cite{liu2022improving}. 
For instance, if $Q$ involves family-related information, but $P_A$ lacks this information, the agent may struggle to provide a relevant response. 
To solve this problem, \citet{liu2022improving} proposed retrieving unseen persona from an external persona pool and examining by an NLI model to ensure consistency between the unseen persona and $P_A$.

Nevertheless, even if the unseen persona is provided, it is difficult for the model to ground on the out-of-distribution persona, i.e., the long-tail problem \cite{liu2022improving}. 
GME \cite{wu-etal-2021-transferable} enforced the grounding on unseen persona $P_{\mathsf{OOD}}$ during inference by masking persona-spans in the original response and re-generate the response conditioned on $P_{\mathsf{OOD}}$, $Q$, and the masked response.
\citet{liu2022improving}, on the other hand, solved this problem by training a persona selection module over $P_A$ and $P_{\mathsf{OOD}}$ (which learned to select $P_{\mathsf{OOD}}$), and generating the response weighted on $P$. 

\begin{table*}[!htp]
\centering
\scriptsize
\resizebox{1.0\linewidth}{!}{
\begin{tabular}{lllllll}
\toprule
\multicolumn{2}{c}{ \textbf{Model}} & \textbf{Fluency} & \textbf{Diversity} & \textbf{Coherence} &\multicolumn{2}{l}{ \textbf{Personalization}} \\\midrule
UA-CVAE &\citeyearpar{lee2022improving} &PPL, ROUGE, METEOR &Dist-1, Dist-2, Dist-3 &UE-Score &\multicolumn{2}{l}{\xmark} \\
BoB &\citeyearpar{song-etal-2021-bob} &PPL &Dist-1, Dist-2, Dist-Avg &\xmark &\multicolumn{2}{l}{Delta Perplexity} \\
PCF &\citeyearpar{wang2023please} &PPL, BLEU &Dist-1, Dist-2, Dist-Avg &UE-Score &\multicolumn{2}{l}{C-Score} \\
LMEDR &\citeyearpar{chen2023learning} &PPL, F1 &Dist-1, Dist-2 &\xmark &\multicolumn{2}{l}{C-Score} \\
SimOAP &\citeyearpar{zhou-etal-2023-simoap} &PPL &Dist-1, Dist-2, Rep &TF-IDF &\multicolumn{2}{l}{C-Score} \\
PAA &\citeyearpar{huang2023personalized} &PPL, BLEU, F1 &Dist-1, Dist-2 &\xmark &\multicolumn{2} {l}{ \xmark} \\
D3 &\citeyearpar{cao-etal-2022-model} &PPL,BLEU,NIST-4, BERTScore &Dist-1, Dist-2, Dist-3, Entropy-n &\xmark &\multicolumn{2} {l}{C-Score} \\
GME &\citeyearpar{wu-etal-2021-transferable} &BLEU &\xmark &\xmark &\multicolumn{2} {l}{C-Score} \\
PS-Transformer &\citeyearpar{liu2022improving} &BLEU, ROUGE, CIDEr &\xmark &\xmark &\multicolumn{2} {l}{ \xmark} \\
DHAP &\citeyearpar{ma2021one} &BLEU, ROUGE, EMB &Dist-1, Dist-2 &\xmark &\multicolumn{2} {l}{Persona-F1, Persona-Coverage} \\
\midrule 
FoCus &\citeyearpar{jang2022call} &PPL, BLEU, ROUGE, chrF++ &\xmark &\xmark &\multicolumn{2} {l}{P-Grounding-Acc} \\
INFO &\citeyearpar{lim-etal-2022-truly} &BLEU, ROUGE, BERTScore, chrF++ &\xmark &\xmark &\multicolumn{2} {l}{P-Grounding-Acc, P-Grounding-F1} \\
WWH &\citeyearpar{kwon-etal-2023-ground} &PPL &\xmark &\xmark &\multicolumn{2} {l}{Persona-F1, Persona-Coverage} \\
IUPD &\citeyearpar{cho-etal-2022-personalized} &PPL &Dist-1, Dist-2 &\xmark &\multicolumn{2} {l}{Persona-Distance} \\
CLV &\citeyearpar{tang-etal-2023-enhancing-personalized} &BLEU, ROUGE &Dist-1, Dist-2 &\multicolumn{2} {l}{Coh-Con-Score} &C-Score* \\
MSP &\citeyearpar{zhong-etal-2022-less} &BLEU, ROUGE, EMB &Dist-1, Dist-2 &\xmark &\multicolumn{2} {l}{Persona-F1, Persona-Coverage} \\
\midrule 
DuLeMon &\citeyearpar{xu-etal-2022-long} &PPL, BLEU, F1 &Dist-1, Dist-2 &\xmark &\multicolumn{2} {l}{\xmark} \\
\bottomrule
\end{tabular}
}
\caption{Evaluation metrics used in each model introduced in Sec.~\ref{sec:approach}. EMB: embedding metric. C-Score*: a different implementation of C-Score.}
\label{tab:evaluation_summary}
\end{table*}

\subsection{Large Language Models and In-Context Learning} 
Given the rising prominence of ChatGPT, we are motivated to examine the impact of large language models (LLMs) and in-context learning on personalized dialogue generation.
As demonstrated in \citet{salewski2023context, jiang2023personallm}, LLMs can reflect personas or personality traits provided in prompts, evident in corresponding personality tests or tasks like writing or reasoning.
Recent works have prompted LLMs as various characters for multi-agent simulation or collaboration \cite{chen2023agentverse,qian2023communicative,Park2023GenerativeAgents,wang2023unleashing}. 
Many works prompted ChatGPT to create synthetic dialogue datasets with distinct personas \cite{lee-etal-2022-personachatgen,chen-etal-2023-places,tu2023characterchat,kim-etal-2024-commonsense}. 
\citet{chen-etal-2023-towards-robust} pretrained their own LLM with persona information augmented to the original dialogue context and demonstrated that prompting such a model improves the agent's persona consistency.

While there's enthusiasm around using prompts with LLMs to imbue them with personas, evaluations typically focus on the success rate of assigned tasks, overlooking the quality of LLM-generated conversations. Furthermore, to the best of our knowledge, existing works have centered on conferring agent personas $P_A$, with no studies exploring using in-context learning to extract or model unknown user personas $P_U$.

\section{Evaluation}
\label{sec:eval}
Personalized dialogue generation literature typically assess the quality of the generated responses across various dimensions. The most commonly examined dimensions include fluency, diversity, coherence, and personalization.

\subsection{Fluency}
\label{sec:eval-fluency}
The fluency evaluation usually refers to the common generation metrics. 
\textit{Perplexity (PPL)} is often reported as an indication of fluency. 
Besides, most works measure the similarity between the generated response to the reference response.
The most popular similarity metrics include the \textbf{lexical overlap metrics}, e.g., \textit{F1}, \textit{BLEU} \cite{papineni2002bleu}, \textit{ROUGE} \cite{lin2004rouge}, \textit{NIST} \cite{doddington2002automatic}, \textit{METEOR} \cite{banerjee-lavie-2005-meteor}, \textit{chrF++} \cite{popovic-2017-chrf}, and \textbf{representation-based metrics}, e.g., \textit{bag-of-word embedding score} \cite{chan-etal-2019-modeling} and \textit{BERTScore} \cite{bert-score}.
Other metric like \textit{CIDEr} \cite{vedantam2015cider} has also been used.

\subsection{Diversity}
\textit{Distinct-1 (Dist-1)} and \textit{Distinct-2 (Dist-2)} \cite{li-etal-2016-diversity} are the most widely adopted metrics for diversity evaluation. 
12 out of 17 papers in Sec.~\ref{sec:approach} includes Dist-1 and Dist-2 in their evaluation.
Entropy-n \cite{zhang2018generating} is the entropy derived from a sentence's n-gram distribution, measuring the uniformity of the empirical n-gram distribution.
\textit{Repetition Rate (Rep)} \cite{zhou-etal-2023-simoap} is proposed for evaluating diversity at the sentence level, which counts the number of identical responses in the candidates that differ from the ground truth.
Comparing the repetition of candidates, this metric is specifically designed for their post-filtering method over the candidate responses. 

\subsection{Coherence}
Coherence refers to the logical and meaningful continuity of a conversation. A Coherent response ensures that the conversation makes sense and flows naturally from the previous turn.
Although coherence evaluation metrics have been proposed \cite{ghazarian-etal-2022-deam,ye-etal-2021-towards-quantifiable}, they were used in none of the papers in Sec~\ref{sec:approach}. 
Many studies omit the coherence evaluation and consider that the Fluency metrics described in the previous section can also measure the coherence of dialogues. 

\citet{zhou-etal-2023-simoap} calculate the cosine similarity between the TF-IDF vectors \cite{salton1988term} of the context and the response as a measure of coherence evaluation.
\textit{Coherence-Consistency Score (Coh-Con.Score)} \cite{tang-etal-2023-enhancing-personalized} measures both dialogue coherence and persona consistency simultaneously.
It takes the query $Q$, response $R$, and persona $P$ as input and assigns 2, 1, 0 for the following scenarios, respectively: both $(P,R)$ and $(Q,R)$ are entailed, only $(P,R)$ is entailed, and otherwise. 

\textit{Utterance Entailment score (UE-Score)} \cite{lee2022improving} compute the NLI score between utterance and response as the coherence score. 

Although both Coh-Con.Score and UE-Score utilize the NLI model, the pretrained data and backbone differ.
While Coh-Con.Score using RoBERTa \cite{liu2019roberta} trained on Dialogue NLI dataset (DNLI) \cite{welleck-etal-2019-dialogue} and finetuned on ConvAI2 and Baidu PersonaChat as the NLI model for English and Chinese, UE-Score finetuned BERT on SNLI dataset \cite{bowman-etal-2015-large}.

\subsection{Personalization}
Personalization can be evaluated from two aspects: consistency and coverage. 

\textbf{Consistency} reflects whether the generated response $R$ is consistent with the given personal information $P$.
\textit{C-Score} \cite{madotto-etal-2019-personalizing} finetuned BERT on DNLI and assigned 1, 0, and -1 for entailment, neutral, and contradiction, respectively. The final C-Score for an utterance is the sum over all personas: $\mathrm{C\mathchar`-Score}(R) = \sum_{i} \mathrm{NLI}(R, P_i)$. 
\textit{P-Score} proposed in \cite{wu-etal-2021-transferable} is actually the same as C-Score. 
\textit{Consistency Score} \cite{tang-etal-2023-enhancing-personalized} reduced the three classes in C-Score to binary classes, i.e., assigning 1 for entailment and neutral labels and 0 for the contradiction label. 

\citet{li-etal-2020-dont} first showed that the perplexity of entailed dialogues would be lower than that of contradicted dialogues. 
Building on this, BoB \cite{song-etal-2021-bob} reported the PPL of entailed and contradicted dialogues, as well as their subtraction, denoted as Delta Perplexity, to highlight the model's capability of distinguishing between entailment and contradiction. 

\textbf{Coverage} shows how much given personal information is reflected in the generated response.
\textit{Persona-F1} \cite{ijcai2019p706} is determined by the overlap of the set of non-stopword unigrams between $R$ and $P$, and \textit{Persona Coverage} \cite{ijcai2019p721} quantifies the IDF-weighted word overlap between $R$ and $P$. 
\textit{Persona Distance} \cite{cho-etal-2022-personalized} is defined as the average word2vec cosine similarity between the keywords of $R$ and $P$, where the keywords are decided by the word frequency after removing stopwords.

\section{Discussion}
\label{sec:discuss}
\subsection{Dataset}
The persona-based datasets are expensive to collect and notable for their limited size. They are also considered artificial as the annotators play the given role and do not act like themselves.
The limited size also causes the out-of-domain persona problem discussed in Sec.~\ref{sec:approach-task-datascarcity}, limiting the adaptation to the real-world scenario.

Crawling from social platforms addresses the issue of data size and provides original utterances from humans. 
However, the quality of these datasets is questionable.
They originate from posts and comments on social media rather than from natural conversations, and thus, the average turns per ``dialogue'' is relatively low.
In addition, the extracted personas may not always align with the actual topic of discussion. Users may also post contradictory statements, making it challenging to deduce a consistent persona. And some situational or fleeting statements might be inadvertently extracted \cite{mazare-etal-2018-training}.

In addition, the diversity of the model are constrained, especially in domain and language variances, as discussed in Sec.~\ref{sec:dataset-domainNlang}.
Multilingual dialogue data is pivotal in capturing the nuances of diverse cultural backgrounds. For example, a model trained solely on English or translated dialogue datasets may not effectively cater to Japanese contexts, given the stark contrast between English's low-context communication style and Japanese's high-context nature.

\subsection{Methodology}
Many works are based on the assumption that the query $Q$, response $R$, and persona $P$ are mutually dependent, that is, assuming the three components are interconnected and that changes or variations in one might influence the other two. 
However, some works overlooked that not all $R$ are personalized, and not all personal information in the given $P$ should be reflected. 

Furthermore, research on personalized dialogue generation is divided into two main streams, and each only models one single party of the dialogue speaker, as described in Sec~\ref{sec:approach-statement}. 
Future research might explore modeling both $P_A$ and $P_U$ for a more comprehensive personalized dialogue generation.

\subsection{Evaluation}
The primary evaluation metrics employed in recent papers may be insufficient. Most works adopted similarity-based metrics for machine translation (MT) or summarization tasks. 
However, the similarity-based metrics have proven ineffective for assessing complex, open-ended tasks such as dialogue generation \cite{gehrmann2023repairing,yeh-etal-2021-comprehensive,deriu2021survey,liu-etal-2016-evaluate}. 
While advanced metrics that align more closely with human judgments have been proposed for dialogue evaluation, none have been applied in the examined studies.

Moreover, each study employs its own data preprocessing and might have different implementation of evaluation methods, complicating direct comparisons between different models. 
We advocate for a standardized approach to data preprocessing and evaluation. Such consistency will enable more precise comparisons among models and ensure that progress in the field is gauged against a uniform benchmark.

\section{Conclusion}
This review delves into personalized dialogue generation, covering Datasets, Methodologies, and Evaluation techniques. The cornerstone dataset in this field is PersonaChat. Modern datasets have expanded in size, added persona grounding, and now cover a broader range of domains, sessions, modalities, and languages. Persona is represented in descriptions, key-value attributes, or basic user IDs to trace past dialogues. The primary methodologies are: (1) imparting a persona to the agent and ensuring its consistency, and (2) pinpointing user personas or choosing the right one for context. The central challenges addressed include maintaining consistency and coherence, appropriately balancing persona with context, choosing pertinent personal details, modeling personas when not directly provided, and navigating limited data. For evaluations, factors like Fluency, Diversity, Coherence, and Personalization are paramount, with perplexity, BLEU, and Distinct-N being commonly used metrics. Personalization assessment primarily gauges persona consistency and coverage.

We conclude by discussing the limitations of the three dimensions. The primary issues with datasets include their limited size, concerns about quality, and insufficient diversity in both domains and languages. Regarding methodology, incorrect assumptions about the interdependencies among persona, query, and response can pose challenges. 
For evaluation, we advocate for consistent assessment standards across various models and benchmarks for effective dialogue evaluation.

We compile information on the publication venues and repositories of related works, including those analyzed in this paper, in Table~\ref{tab:exhaustive_list} in the Appendix. 

\section{Acknowledgements}
This work was supported by the Institute of AI and Beyond of the University of Tokyo and MEXT KAKENHI Grant Number JP22H05015. Chen is also supported by JST SPRING, Grant Number JPMJSP2108.

\nocite{*}
\section{Bibliographical References}\label{sec:reference}

\bibliographystyle{lrec-coling2024-natbib}
\bibliography{custom}


\section{Appendix}
\label{paper_list}
Table~\ref{tab:exhaustive_list} presents a list of works related to personalized dialogue generation with their publication venues and repositories.

\begin{sidewaystable*}[h]
\centering
\scriptsize
\resizebox{1.0\linewidth}{!}{
\begin{tabular}{llllll}
\toprule
Dataset & Model & Title &  & Venue & Repo \\
\midrule
&BoB &BERT Over BERT for Training Persona-based Dialogue Models from Limited Personalized Data &\citet{song-etal-2021-bob} &ACL &\href{https://github.com/songhaoyu/BoB}{link} \\
&PABST &Unsupervised Enrichment of Persona-grounded Dialog with Background Stories &\citet{majumder-etal-2021-unsupervised} &ACL &\href{https://github.com/majumderb/pabst}{link} \\
PersonaMinEdit &GME &Transferable Persona-Grounded Dialogues via Grounded Minimal Edits &\citet{wu-etal-2021-transferable} &EMNLP &\href{https://github.com/thu-coai/grounded-minimal-edit}{link} \\
PER-CHAT & &Personalized Response Generation via Generative Split Memory Network &\citet{wu-etal-2021-personalized} &NAACL &\href{https://github.com/Willyoung2017/PER-CHAT}{link} \\
Persona Reddit & &A Simple and Efficient Multi-Task Learning Approach for Conditioned Dialogue Generation &\citet{zeng-nie-2021-simple} &NAACL &\href{https://github.com/zengyan-97/MultiT-C-Dialog}{link} \\
&DHAP &One Chatbot Per Person: Creating Personalized Chatbots based on Implicit User Profiles &\citet{ma2021one} &SIGIR &\href{https://github.com/zhengyima/DHAP}{link} \\
Pchatbot{W/L} & &Pchatbot: A Large-Scale Dataset for Personalized Chatbot &\citet{qian2021pchatbot} &SIGIR &\href{https://github.com/qhjqhj00/SIGIR2021-Pchatbot}{link} \\
XPersona & &XPersona: Evaluating Multilingual Personalized Chatbot &\citet{lin2021xpersona} &NLP4ConvAI  &\href{https://github.com/HLTCHKUST/Xpersona}{link} \\
FoCus &FoCus &Call for Customized Conversation: Customized Conversation Grounding Persona and Knowledge &\citet{jang2022call} &AAAI &\href{https://github.com/pkchat-focus/FoCus}{link} \\
& &Dual Task Framework for Improving Persona-grounded Dialogue Dataset &\citet{kim2022dual} &AAAI &x \\
&D3 &A Model-Agnostic Data Manipulation Method for Persona-based Dialogue Generation &\citet{cao-etal-2022-model} &ACL &\href{https://github.com/caoyu-noob/D3}{link} \\
& &DialogVED: A Pre-trained Latent Variable Encoder-Decoder Model for Dialog Response Generation &\citet{chen-etal-2022-dialogved} &ACL &\href{https://github.com/lemuria-wchen/DialogVED}{link} \\
MSC &MSC &Beyond Goldfish Memory : Long-Term Open-Domain Conversation &\citet{xu-etal-2022-beyond} &ACL &\href{https://parl.ai/projects/msc/}{link} \\
DuLeMon &DuLeMon &Long Time No See! Open-Domain Conversation with Long-Term Persona Memory &\citet{xu-etal-2022-long} &ACL-Findings &\href{https://github.com/PaddlePaddle/Research/tree/master/NLP/ACL2022-DuLeMon}{link} \\
IT-ConvAI2 &PS-Transformer &Improving Personality Consistency in Conversation by Persona Extending &\citet{liu2022improving} &CIKM &\href{https://github.com/CCIIPLab/Persona\_Extend}{link} \\
&IUPD &A Personalized Dialogue Generator with Implicit User Persona Detection &\citet{cho-etal-2022-personalized} &COLING &x \\
&INFO &You Truly Understand What I Need: Intellectual and Friendly Dialogue Agents grounding Knowledge and Persona &\citet{lim-etal-2022-truly} &EMNLP-Findings &x \\
& &Keep Me Updated! Memory Management in Long-term Conversations &\citet{bae-etal-2022-keep} &EMNLP-Findings &\href{https://github.com/naver-ai/carecall-memory}{link} \\
&UA-CVAE &Improving Contextual Coherence in Variational Personalized and Empathetic Dialogue Agents &\citet{lee2022improving} &ICASSP &x \\
&MSP &Less is More: Learning to Refine Dialogue History for Personalized Dialogue Generation &\citet{zhong-etal-2022-less} &NAACL &\href{https://github.com/bangbangbang12315/MSP}{link} \\
& &You Don't Know My Favorite Color: Preventing Dialogue Representations from Revealing Speakers' Private Personas &\citet{li-etal-2022-dont} &NAACL &\href{https://github.com/hkust-knowcomp/persona\_leakage\_and\_defense\_in\_gpt-2}{link} \\
&LMEDR &Learning to Memorize Entailment and Discourse Relations for Persona-Consistent Dialogues &\citet{chen2023learning} &AAAI &\href{https://github.com/Chenrj233/LMEDR}{link} \\
&PAA &Personalized Dialogue Generation with Persona-Adaptive Attention &\citet{huang2023personalized} &AAAI &\href{https://github.com/hqsiswiliam/persona-adaptive-attention}{link} \\
&CLV &Enhancing Personalized Dialogue Generation with Contrastive Latent Variables: Combining Sparse and Dense Persona &\citet{tang-etal-2023-enhancing-personalized} &ACL &\href{https://github.com/toyhom/clv}{link} \\
&SimOAP &SimOAP: Improve Coherence and Consistency in Persona-based Dialogue Generation via Over-sampling and Post-evaluation &\citet{zhou-etal-2023-simoap} &ACL &x \\
MPChat & &MPCHAT- Towards Multimodal Persona-Grounded Conversation &\citet{ahn-etal-2023-mpchat} &ACL &\href{https://github.com/ahnjaewoo/MPCHAT}{link} \\
LiveChat & &LiveChat: A Large-Scale Personalized Dialogue Dataset Automatically Constructed from Live Streaming &\citet{gao-etal-2023-livechat} &ACL &\href{https://github.com/gaojingsheng/LiveChat}{link} \\
& &RECAP: Retrieval-Enhanced Context-Aware Prefix Encoder for Personalized Dialogue Response Generation &\citet{liu-etal-2023-recap} &ACL &\href{https://github.com/isi-nlp/recap}{link} \\
& &PAED: Zero-Shot Persona Attribute Extraction in Dialogues &\citet{zhu-etal-2023-paed} &ACL &\href{https://github.com/SenticNet/PAED}{link} \\
& &Multimodal Persona Based Generation of Comic Dialogs &\citet{agrawal-etal-2023-multimodal} &ACL &\href{https://github.com/Atenrev/comics-dialogue-generation}{link} \\
& &Towards Zero-Shot Persona Dialogue Generation with In-Context Learning &\citet{xu-etal-2023-towards-zero} &ACL-Findings &x \\
& &PAL: Persona-Augmented Emotional Support Conversation Generation &\citet{cheng-etal-2023-pal} &ACL-Findings &\href{https://github.com/chengjl19/PAL}{link} \\
& &Learning to Predict Persona Information for Dialogue Personalization without Explicit Persona Description &\citet{zhou-etal-2023-learning} &ACL-Findings &x \\
& &Towards Robust Personalized Dialogue Generation via Order-Insensitive Representation Regularization &\citet{chen-etal-2023-towards-robust} &ACL-Findings &\href{https://github.com/ChanLiang/ORIG}{link} \\
MSPD &WWH &WHAT, WHEN, and HOW to Ground: Designing User Persona-Aware Conversational Agents for Engaging Dialogue &\citet{kwon-etal-2023-ground} &ACL-Industry &x \\
& &Personalized Quest and Dialogue Generation in Role-Playing Games: A Knowledge Graph- and Language Model-based Approach &\citet{ashby2023personalized} &CHI &\href{https://github.com/DRAGNLabs/DRAGN-Town-Quests}{link} \\
&PCF &Please don't answer out of context: Personalized Dialogue Generation Fusing Persona and Context &\citet{wang2023please} &IJCNN &x \\
\bottomrule
\end{tabular}
}
\caption{List of works related to personalized dialogue generation. The first two columns are the names used in Sec~\ref{sec:dataset} and ~\ref{sec:approach}.}
\label{tab:exhaustive_list}
\end{sidewaystable*}

\end{document}